\documentclass[journal]{IEEEtran}

\ifCLASSINFOpdf
\else
   \usepackage[dvips]{graphicx}
\fi
\usepackage{url}

\usepackage{amsmath,amssymb,amsfonts}
\usepackage{algorithmic}
\usepackage{graphicx}
\usepackage{textcomp}
\usepackage{xcolor}
\usepackage{booktabs}
\usepackage{cleveref}

\usepackage[style=ieee, backend=biber, mincitenames=1, maxcitenames=2]{biblatex}
\begin{document}

\title{
UNet-AF:
Alias-free UNet architectures
}

\author{Jérémy Scanvic, Quentin Barthélemy, Juli\'an Tachella
\thanks{This project was provided with computing HPC and storage resources by GENCI at IDRIS thanks to the grant 2025-AD011016422 on the supercomputer Jean Zay’s H100 partition.}
\thanks{Jérémy Scanvic (jeremy.scanvic@ens-lyon.fr) and Juli\'an Tachella (julian.tachella@ens-lyon.fr) are with ENSL, CNRS, Laboratoire de Physique, F-69364 Lyon, France. Jérémy Scanvic and Quentin Barthélemy (q.barthelemy@hirschsecure.fr) are with Hirsch Secure, Lyon, France.}
}

\markboth{}{}

\maketitle

\begin{abstract}
The simplicity and effectiveness of UNet architectures make them ubiquitous in image restoration,
segmentation,
and diffusion models.
They are often assumed to be equivariant to translations,
yet they traditionally consist of layers that are known to be prone to aliasing, which hinders their equivariance in practice.
To overcome this limitation,
we show how to build sub-pixel translation-equivariant UNet architectures by appropriately choosing their main components (convolution, pooling, downsampling, activation, and normalization layers) to be alias-free.
We evaluate the proposed equivariant architectures against non-equivariant baselines
on image restoration tasks and observe competitive performance with a significant increase in measured equivariance.
Through extensive ablation studies,
we also demonstrate the importance of every architectural choice to achieve high equivariance\footnote{Our implementation is available at \url{https://github.com/jscanvic/UNet-AF}}.
\end{abstract}


\IEEEpeerreviewmaketitle

\section{Introduction}

UNet networks are encoder-decoder neural networks that operate across multiple scales using pairs of pooling/downsampling and upsampling layers, and have concatenation layers bridging the encoder and decoder at each scale.
Notably, they are commonly applied in medical imaging~\cite{ronneberger15UNet,jin17Deep}, in image super-resolution~\cite{zhang20Deep},
and in diffusion models~\cite{crowson24Scalable}.

In many applications, UNet networks are expected to be equivariant to translations, i.e., that their output translates with their input.
Convolutional neural network (CNN) layers~\cite{lecun98Gradientbased} are often preferred to promote translation-equivariance,
however recent work
shows that standard CNNs only satisfy it approximately,
mostly due to
aliasing in pooling and upsampling layers~\cite{azulay19Why,zhang19Making,zou20Delving}.
The authors propose to combine pooling and upsampling layers with anti-aliasing filters to reduce aliasing, and thus increase approximate equivariance.

\textcite{chaman21TrulyEquivariant} identify activation functions as another source of aliasing, which hinders
translation-equivariance.
They introduce adaptive pooling and upsampling
whereby the sampling grid shifts with the input,
thereby recovering perfect equivariance to whole-pixel translations.
However, this solution does not improve equivariance to sub-pixel (continuous) translations~\cite{michaeli23AliasFree}.

\begin{figure}[thbp]
    \centering
    \includegraphics{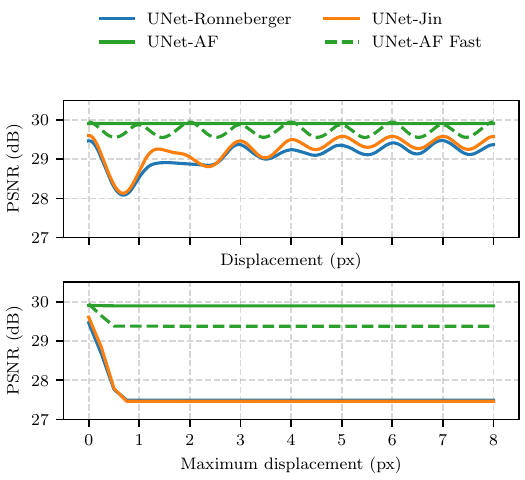}
    \caption{
    \textbf{
    Equivariance and robustness to circular translations.
    }
    (Top) Performance on an horizontally translated image, with 0.01\,px steps, (bottom) adversarial performance on the worst translation up to a maximum displacement, with 0.25\,px steps.
    UNet-AF is stable and robust, unlike the baselines.
    }
    \label{fig:equiv}
\end{figure}

\textcite{karras21AliasFree} emphasize the importance of sub-pixel equivariance for image generation to avoid texture sticking, and reframe the problem
using a traditional approach built upon sampling theory.
Discrete finite images are viewed as representations of bandlimited periodic images related through sampling and Whittaker-Shannon (sinc) interpolation~\cite{vetterli14Foundations}.
Instead of using adaptive sampling, they reduce the aliasing caused by activation functions by low-pass filtering them.
They also advocate for the use of anti-aliasing filters with a sharper cut-off, and a flatter response in the pass and stop bands, although more computationally expensive.

\textcite{michaeli23AliasFree} go one step further and use exact sinc filters implemented using the fast Fourier transform~(FFT).
They identify the lack of smoothness of ReLU as another cause of aliasing and replace it with polynomial activations, and also propose an alias-free substitute to layer normalization.

\begin{figure*}[thbp]
    \centering
    \includegraphics{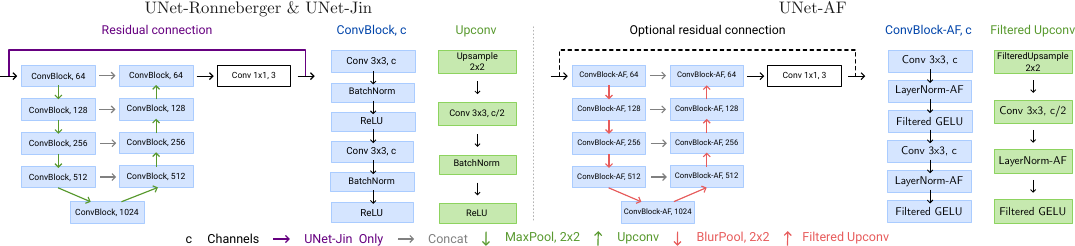}
    \caption{\textbf{Overview of the architecture.} UNet-AF is obtained by changing the architectures of~\textcite{ronneberger15UNet}, and~\textcite{jin17Deep}. Anti-aliasing filters are added in upsampling layers, max-pooling layers are replaced with blur pooling layers, ReLU activations are replaced with filtered GELU layers and batch normalization is replaced by alias-free layer normalization.
    The padding mode and residual connection are specified on a per-task basis.
    }
    \label{fig:arch}
\end{figure*}

While the design of sub-pixel translation-equivariant architectures has been successfully explored for classifiers~(encoder type)~\cite{michaeli23AliasFree,michaeli25AliasFree}, and image generators (decoder type)~\cite{karras21AliasFree},
it is, to the best of our knowledge, not the case for UNet architectures (encoder-decoder with skip-connections).

Our contributions are the following:

\begin{itemize}
    \item We propose a recipe for building alias-free UNet architectures.
    \item We evaluate its effectiveness, training stability, computational efficiency, and empirical equivariance on multiple image restoration tasks (see summary in~\Cref{fig:equiv}).
    \item We conduct extensive ablation studies to validate its design, and identify its tradeoffs.
\end{itemize}

\section{Alias-free UNet} \label{sec:arch}

In this section, we discuss every type of layer used in typical UNet architectures, i.e., convolutions, attention layers, residual connections, concatenation layers, pooling layers, upsampling layers, activation functions, and normalization layers. In each case, we present the necessary changes to obtain a translation-equivariant UNet architecture.

We follow the approach of~\textcite{michaeli23AliasFree}, and model sub-pixel translation of discrete images $x \in \mathbb{R}^{H \times W}\ (H, W \geq 1)$ as resampling on a translated grid using the periodic sinc interpolant~\cite{birdsall88Minimum}.
Unlike other translation schemes relying on non-circular padding (zeros, reflection), or non-ideal interpolants (bilinear, bicubic), periodic sinc translation is invertible~\cite{vetterli14Foundations}, which makes exact equivariance possible.

Formally, we denote as $\mathop{T_g} x$ the translation of $x$ by displacement $g \in [0, H) \times [0, W)$ where $(1, 0)$ and $(0, 1)$ denote respectively a downwards and rightwards translation of 1\,px.
See the appendix for a closed-form formula of it. In this setting, translation-equivariance is expressed for a network $f(x)$ by the equation
\begin{equation} \label{eq:equiv}
    f(\mathop{T_g} x) = \mathop{T_g} f(x), \ \forall x \in \mathbb R^{H \times W},  \forall g \in [0, H) \times [0, W),
\end{equation}
that is, translating the input results in a translated output.

We focus on equivariance to circular translations as it has been shown to generalize well to other translations, notably crop-translations~\cite{michaeli23AliasFree}. We suggest replacements for equivariance to these translations specifically when it does not hinder performance substantially.

\paragraph{Convolutions}

While convolutions are not prone to aliasing, a mismatch in padding mode with the translations causes boundary errors, and thus a loss of equivariance.
Therefore, we use circular convolutions for circular translations, and zero-padded convolutions for crop-translations~\cite{zhang19Making}. These convolutions are finally arranged in blocks (ConvBlock-AF)
similarly to
the ones in the base architecture (ConvBlock).
Attention mechanisms can also be incorporated in the form of channel attention~\cite{ali21xcit}, which is translation-equivariant~\cite{michaeli25AliasFree}.

\paragraph{Residual connections}

The global residual connection featured in certain UNet architectures~\cite{jin17Deep} does not introduce aliasing and can be retained.

\paragraph{Concatenation layers}

Similarly to residual connections, the intermediate concatenations found in most UNet architectures do not introduce aliasing and can also be retained.

\paragraph{Pooling layers}

We substitute max-pooling for blur pooling with periodic sinc anti-aliasing implemented using the FFT, similarly to~\textcite{michaeli23AliasFree}.
Due to the contribution of every input pixel to each output pixel, this anti-aliasing scheme is more computationally costly than the filters used by~\textcite{zhang19Making,chaman21TrulyEquivariant,karras21AliasFree}, including the box filter used in average pooling, and windowed sinc filtering.
Instead, it eliminates aliasing within numerical precision, which results in higher equivariance.
We use this anti-aliasing consistently in other layers.

\paragraph{Upsampling layers}

The original upsampling scheme consists in merely inserting zeros between input pixels,
which preserves input information in the output low-frequencies, yet fills the high-frequencies with aliasing.
It is eliminated in the proposed architecture using an anti-aliasing filter.

\paragraph{Activation functions}

We replace ReLU activations with filtered activation functions~\cite{karras21AliasFree,michaeli23AliasFree}, which consist in first resampling the input on a finer grid, applying a base activation function, removing the high-frequencies, and resampling on the coarse grid.
Part of the aliasing is removed with the high-frequencies, thereby improving equivariance. The upsampling rate is set to two, following previous work.

The GELU function is chosen as the underlying base activation function, similarly to~\textcite{michaeli25AliasFree}.
Its smoothness limits the addition of high-frequencies in the finest grid, which improves equivariance, and unlike polynomial activations~\cite{michaeli23AliasFree}, its linear growth prevents it from amplifying numerical equivariance error.

Filtered activations incur an additional computational cost compared to standard activations~\cite{karras21AliasFree}. Here, we propose a variant that achieves better equivariance metrics than that of~\textcite{karras21AliasFree}, and is faster than that of~\textcite{michaeli23AliasFree}.
This variant (UNet-AF Fast) is detailed in the appendix.

\paragraph{Normalization layers}

\textcite{scanvic25ScaleEquivariant} show theo\-retically and empirically that batch normalization, instance normalization, and alias-free layer-normalization~\cite{michaeli23AliasFree} are equivariant to translations, but standard layer normalization~\cite{liu22ConvNet} is not. Among these three equivariant options, we recommend using alias-free layer normalization for its higher training stability and performance.

\begin{table*}[hbtp]
    \centering
    \caption{
    \textbf{Performance on circular deblurring.} The changes in~\Cref{sec:arch} lead to a steady increase in equivariance (EQUIV), stabilize the training dynamics and improve performance (PSNR, SSIM, LPIPS), at the expense of a higher computational cost (FPS). \\
    Best metric in \textbf{bold}, second best \underline{underlined}.
    Values: avg ± st.d.
    }
    \label{tbl:main}
    \setlength{\tabcolsep}{9pt}
    \begin{tabular}{llcccccc}
        \toprule
        Method & Difference & PSNR $\uparrow$ & SSIM $\uparrow$ & LPIPS $\downarrow$ & EQUIV $\uparrow$ & STAB $\uparrow$ & FPS $\uparrow$ \\
        \midrule
        UNet-Ronneberger & & 31.28 ± 2.13 & 0.9316 ± 0.0271 & 0.0377 ± 0.0235 & 37.60 ± 2.62 & 9.10 & \textbf{2494} \\
        UNet-Jin & + Res conn. & 31.64 ± 2.51 & 0.9280 ± 0.0345 & 0.0402 ± 0.0254 & 38.20 ± 2.22 & 12.86 & \underline{2114} \\
        & + LayerNorm-AF & 32.16 ± 2.37 & 0.9383 ± 0.0211 & 0.0355 ± 0.0210 & 38.41 ± 1.99 & 18.40 & 1230 \\
        & + Circular pad. & \textbf{32.46 ± 2.45} & \textbf{0.9417 ± 0.0209} & \underline{0.0323 ± 0.0199} & 43.16 ± 2.38 & 20.63 & 982 \\
        & + Filtered GELU & 32.22 ± 2.39 & 0.9383 ± 0.0201 & 0.0331 ± 0.0193 & 43.55 ± 2.77 & 20.48 & 302 \\
        & + Filtered Upconv. & 32.24 ± 2.35 & 0.9382 ± 0.0201 & 0.0333 ± 0.0197 & 46.33 ± 2.59 & 20.34 & 293 \\
        UNet-AF (Ours) & + BlurPool & 32.22 ± 2.32 & 0.9373 ± 0.0204 & 0.0330 ± 0.0210 & \textbf{81.63 ± 0.26} & \underline{20.81} & 290 \\
        UNet-AF Fast (Ours) & + FIR Filtered ReLU & \underline{32.39 ± 2.40} & \underline{0.9408 ± 0.0207} & \textbf{0.0307 ± 0.0181} & 49.36 ± 2.84 & \textbf{23.29} & 552 \\
        \midrule
        \multicolumn{2}{l}{Blurred images} & 26.24 ± 2.30 & 0.8170 ± 0.0406 & 0.2258 ± 0.0719 & N/A & N/A & N/A \\
        \bottomrule
    \end{tabular}
\end{table*}

\section{Experiments}

We apply the changes described in~\Cref{sec:arch} to the UNet architecture
of~\textcite{jin17Deep}~(UNet-Jin), which features an additional residual connection on top of
the original UNet architecture of~\textcite{ronneberger15UNet}~(UNet-Ronneberger).
The resulting architectures (UNet-AF, and UNet-AF Fast) render equivariant these non-equivariant baselines.
While we focus on these specific UNet architectures, our analysis can be easily applied to other UNet backbones.
The architecture of UNet-AF is represented in~\Cref{fig:arch},
and we conduct extensive experiments to evaluate it and UNet-AF Fast.

We train
the base architectures and equivariant counter-parts
to solve imaging inverse problems, and we compare the resulting models using various metrics.
In~\Cref{subsec:restoration}, we describe the different tasks in more detail, and we compare the performance of the three architectures, and in~\Cref{subsec:ablations} we report extensive ablation studies on the proposed architecture.

For a fair comparison, the different models are trained using the same training procedure from the DeepInverse library~\cite{tachella25DeepInverse}.
It minimizes a mean squared error loss using Adam~\cite{kingma17Adam} with a learning rate of 5e-4, a batch size of 5, and no weight decay.

The architectures are evaluated using two distortion metrics, the PSNR and SSIM~\cite{wang04Image}, one perceptual metric, LPIPS~\cite{zhang18Unreasonable}, one equivariance metric, EQUIV~\cite{sechaud26Equivariant}, one training stability metric, STAB, and one metric of computational efficiency, FPS.
Precisely, EQUIV is the average PSNR of the equivariance error associated to~\cref{eq:equiv}, STAB is the variance in dB of the evaluation PSNR increments from one epoch to the next, and FPS is the average number of images processed in a second at inference, measured on an NVIDIA H100 GPU.

\subsection{Image restoration} \label{subsec:restoration}

Three image restoration tasks are considered,
1) image deblurring with circular boundary conditions, 2) deblurring with cropped boundaries, and 3) denoising.
We pre-process the 900 natural images from DIV2K~\cite{agustsson17NTIRE} to constitute the ground truths, which we further split into 700 training images, 100 evaluation images, and 100 test images.
Each image is first resized to 256\,px and then its center 128\,px are cropped so that information at the boundary is available to compute crop-translations similarly to~\textcite{zhang19Making}.

Image restoration consists in recovering high-quality images $x_{\text{HQ}}$ from degraded images $x_{\text{LQ}}$. In our experiments, we consider settings with blur and noise
\begin{equation}
    x_{\text{LQ}} = h * x_{\text{HQ}} + \varepsilon
\end{equation}
where $h$ denotes a Gaussian blur kernel and $\varepsilon$ is additive white Gaussian noise.
For tasks 1) and 2), we use a Gaussian kernel with standard deviation 1\,px and Gaussian noise with a standard deviation of $0.01$, and for task 3) we use Gaussian noise with a standard deviation of $0.1$ and no blur.

We train each network to directly map degraded images to high-quality images.
As long as the degradation process is equivariant to translations, which is the case for uniform blur and additive white Gaussian noise,
it suffices that the network is equivariant to render the whole system equivariant~\cite{chen23Imaging,sechaud26Equivariant}
\begin{equation}
    f\left(h * (\mathop{T_g} x_{\text{HQ}}\right) + \mathop{T_g} \varepsilon) = \mathop{T_g} f(h * x_{\text{HQ}} + \varepsilon).
\end{equation}

We use different variants of UNet-AF depending on the specific task. For circular deblurring and denoising, we use circular padding, and for valid deblurring, we use zero-padding. Moreover, the residual connection is present for deblurring tasks, but not for denoising, for improved performance and training stability.

\Cref{tbl:main} shows that the
changes in~\Cref{sec:arch} result in a steady increase in equivariance from the baseline to UNet-AF, with a final difference of about 40\,dB.
Moreover,
our architecture is both more performant and more stable at training than the two baselines.
The variant UNet-AF Fast has twice the throughput of UNet-AF at the expense of sitting between it and the baselines in equivariance.
Different reconstructions are shown in the appendix.

Unlike the experiments of~\textcite{michaeli23AliasFree} on alias-free classifiers, we observe no significant drop in performance due to blur pooling, and neither for alias-free layer normalization which we find to significantly increase performance as well as training stability.
However, we observe a slight drop in performance due to the filtered activation functions as well, and we believe that increasing the depth of the network might compensate for this change.

Inference speed is mostly hindered by the change in normalization and activation functions, with the final equivariant network being about 7 times slower than the starting implementation of UNet-Jin.
Replacing standard layers with alias-free alternatives is also detrimental to computational efficiency, which is explained by the presence of additional filters.

\begin{figure}[b!]
    \centering
    \includegraphics{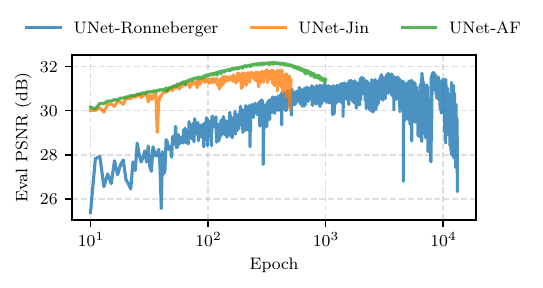}
    \caption{
    \textbf{Evaluation PSNR throughout training for circular deblurring.}
    Our proposed model has significantly smoother training dynamics than the non-equivariant baselines.
    }
    \label{fig:training_curves}
\end{figure}

The difference in training stability is illustrated in~\Cref{fig:training_curves}, which shows the validation PSNR for each epoch.
Moreover, the difference in equivariance is illustrated in~\Cref{fig:equiv}, which shows the PSNR for an image translated by varying displacements on average (top), or adversarially (bottom).
The noticeable phase change around 2\,px
is explained by the transition from an equivariance error dominated by boundary artifacts that only affect small translations, to a regime where the smaller aliasing error that is always present dominates.

Results on additional tasks are presented in the appendix.

\begin{table}[b!]
    \centering
    \caption{
    \textbf{Ablation study.} Each design choice made in~\Cref{sec:arch} is explored in terms of performance and equivariance.
    Best metric in \textbf{bold}, second best \underline{underlined}.
    Values: avg ± st.d.
    }
    \label{tbl:ablation}
    \setlength{\tabcolsep}{8pt}
    \begin{tabular}{llcc}
        \toprule
    Replacement & Variant & PSNR $\uparrow$ & EQUIV $\uparrow$ \\
	\midrule
	  Res. connection & Present & \textbf{32.22 ± 2.32} & \textbf{81.63 ± 0.26} \\
    & Absent & \underline{31.96 ± 2.33} & \underline{77.77 ± 2.32} \\
	\midrule
	Normalization & LayerNorm-AF & \underline{32.22 ± 2.32} & \underline{81.63 ± 0.26} \\
    & BatchNorm & 32.21 ± 2.31 & 80.62 ± 8.14 \\
    & InstanceNorm & 31.50 ± 2.19 & 78.40 ± 0.42 \\
    & LayerNorm & \textbf{32.31 ± 2.36} & 54.19 ± 3.03 \\
    & None & 32.10 ± 2.39 & \textbf{84.78 ± 3.91} \\
    \midrule
	Padding & Circular & \textbf{32.22 ± 2.32} & \textbf{81.63 ± 0.26} \\
    & Zeros & \underline{32.14 ± 2.36} & \underline{40.35 ± 1.95} \\
    & Reflect & 32.09 ± 2.38 & 40.24 ± 2.10 \\
	\midrule
	Activation & Filtered GELU & 32.22 ± 2.32 & \textbf{81.63 ± 0.26} \\
    & GELU & \underline{32.28 ± 2.41} & 47.03 ± 2.83 \\
    & Filtered ReLU & \textbf{32.28 ± 2.35} & \underline{78.95 ± 1.21} \\
    & ReLU & 32.27 ± 2.42 & 45.38 ± 2.80 \\
    & Filtered Poly & 26.29 ± 2.20 & 55.56 ± 0.86 \\
    & Poly & 26.25 ± 2.25 & 53.96 ± 0.95 \\
	\midrule
    Upsampling & Filtered & \textbf{32.22 ± 2.32} & \textbf{81.63 ± 0.26} \\
   	& Unfiltered & \underline{32.15 ± 2.33} & \underline{45.90 ± 2.69} \\
	\midrule
	Pooling & BlurPool & 32.22 ± 2.32 & \textbf{81.63 ± 0.26} \\
	& MaxPool & \underline{32.24 ± 2.35} & 46.33 ± 2.59 \\
    & AvgPool & 32.15 ± 2.36 & 46.59 ± 2.71 \\
    & MaxBlurPool & \textbf{32.32 ± 2.40} & \underline{47.03 ± 3.10} \\
    \bottomrule
    \end{tabular}
\end{table}

\subsection{Ablation studies} \label{subsec:ablations}

We assess the impact of each choice described in~\Cref{sec:arch}.
Specifically,  each model is trained on the circular deblurring task introduced in~\Cref{subsec:restoration}, and \Cref{tbl:ablation} shows the results for each layer, including residual connections, normalization layers, padding in convolutions, activation functions, and upsampling and pooling layers.

We observe a drop of about 40\,dB in equivariance when any of the layers is replaced by an alternative prone to aliasing, thereby showing that each change in \Cref{sec:arch} equally contributes to the equivariance of the final model.
In this setting, we also observe a noticeable increase in performance for the variant with a residual connection.

Despite being theoretically free of aliasing, we find that filtered polynomial activations measurably hinder equivariance.
We believe that this error stems from the amplification of prior numerical error when it passes in the quadratic terms of activation layers.
Moreover, unlike~\textcite{michaeli23AliasFree} but similarly to~\textcite{michaeli25AliasFree}, we find that these layers result in a significant decrease in performance.
As explained in \Cref{sec:arch}, the use of filtered GELU instead of filtered ReLU as the base activation layer leads to slightly better equivariance.

We find that despite averaging neighboring pixel values, average pooling hinders equivariance as much as regular max pooling,
which is consistent with the relatively low attenuation of box filters in the stop band.
Moreover, despite having been proposed by~\textcite{zhang19Making} to improve the equivariance of CNNs, we find that MaxBlurPool is as poor as max pooling and blur pooling in terms of equivariance.

We find that, consistently with previous work~\cite{scanvic25Translationequivariance}, layer normalization is detrimental to equivariance, but that instance normalization, batch normalization and alias-free layer normalization are not.
Moreover, we observe that instance normalization performs significantly worse than batch normalization and alias-free layer normalization, and that alias-free layer normalization performs slightly better than batch normalization.
We also observe that removing the normalization layers altogether leads to only a small gain in equivariance, despite removing all of the numerical error they introduce, which shows that alias-free layer normalization is numerically close to optimal.

\section{Conclusion}

We propose a recipe to build sub-pixel translation-equivariant UNet architectures
using a careful selection of state-of-the-art equivariant layers.
Moreover, we propose fast alternatives that offer a good trade-off between equivariance and computational complexity.
We validate our approach extensively on image restoration tasks including circular and valid Gaussian deblurring, and Gaussian denoising.
We also conduct extensive ablation studies to validate our design choices and find that each of the substitutions we make is necessary for the overall equivariance of the architecture.
Our experiments show that equivariance to translations tends to be correlated with performance and training stability, but at a greater computational expense.

\renewcommand*{\bibfont}{\footnotesize}
\printbibliography

\cleardoublepage

\appendix

In~\Cref{sec:method:supp}, we give additional details about sub-pixel translation-equivariance, and in~\Cref{sec:experiments:supp}, we give additional details about the experiments.

\subsection{Equivariance to translations} \label{sec:method:supp}

In~\Cref{subsec:continuous_translations}, we express the formula of the continuous periodic translations used in the equivariance metric, and in~\Cref{subsec:fast_activations}, we present the fast filtered activation functions that we use in the faster variant of our proposed architecture (UNet-AF Fast).

\subsubsection{Continuous translations} \label{subsec:continuous_translations}

The continuous periodic translation used in the equivariance metric is defined from the ideal periodic sinc interpolant~\cite{birdsall88Minimum}:
\begin{equation}
    \mathrm{psinc}(t) = \begin{cases}
        \frac{\sin(\pi t)}{N \sin(\pi t / N)} &\text{if $N$ is odd,} \\
        \frac{\sin(\pi t)}{N \tan(\pi t / N)} &\text{if $N$ is even,}
    \end{cases}
\end{equation}
where $t \in [0, N)$. We express it in one dimension for the sake of simplicity, and extend it to two dimensions using the separability of horizontal and vertical directions.
Precisely, for displacement $g \in \mathbb [0, N)$, and $x \in \mathbb R^N$, it is defined by
\begin{equation}
    (\mathop{T_g} x)_{k} = \sum_{l=0}^{N-1} \mathrm{psinc}(k - g - l) x_l.
\end{equation}

\subsubsection{Fast filtered activations} \label{subsec:fast_activations}

We propose a faster variant treading the equivariance-efficiency trade-off. It is obtained by a series of changes made to the activation functions. First, the base GELU activation is replaced by a ReLU activation, and the FFT-based filtering is changed to convolution-based filtering. These two substitutions are made to enable the use of the efficient hardware accelerated implementation from~\textcite{karras21AliasFree}.

However, it does not support circular convolutions, and results in a loss of equivariance due to boundary conditions.
To overcome this,
one of our contributions is to extend this implementation to support circular convolutions.

Instead of using Kaiser-windowed sinc filters with manually tuned window parameters similarly to~\textcite{karras21AliasFree}, or unwindowed periodic sinc filters similarly to~\textcite{michaeli23AliasFree}, we tune multiple 13-taps filters using a data-driven approach.

Precisely, we select an arbitrary input image from the dataset and we select filters that minimize the mean squared error between the output of the input image when passed through the ideal unwindowed periodic sinc activation, and the output of the approximated activation using less taps. For this, we use the L-BFGS optimizer featured in PyTorch with a learning rate of 1.0 and a maximum of 200 steps.

In total, we tune 10 different filters clustered in 5 pairs for the 5 scales in our architecture, with each pair consisting of the filter following the upsampling step, and that preceding the downsampling step.

\subsection{Experiments details} \label{sec:experiments:supp}

\begin{figure*}[tbhp]
    \centering
    \includegraphics{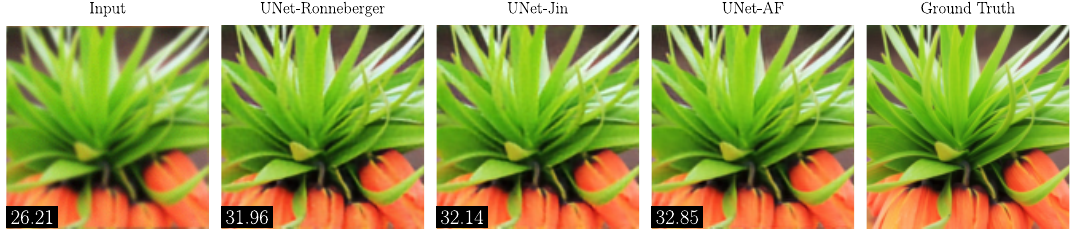}
    \caption{
    \textbf{Reconstructions of a circularly blurred image.} The PSNR is displayed in the bottom-left corner of each image.
    }
    \label{fig:recons}
\end{figure*}

In~\Cref{subsec:additional_tasks}, we present results on additional tasks, in~\Cref{subsec:ablation_unet_af_fast}, we present an ablation on the design choices made in the fast variant of our proposed design (UNet-AF Fast), and \Cref{fig:recons} shows examples of reconstructions for the main deblurring task.

\subsubsection{Additional tasks} \label{subsec:additional_tasks}

\Cref{tbl:additional_tasks} shows the results on valid deblurring, and we observe that UNet-AF is slightly more equivariant to crop-translations than UNet-Ronneberger and UNet-Jin. Unlike circular translations, these are not invertible as they discard and introduce information at the boundaries, and it is expected that the equivariance errors are larger.
Moreover, UNet-AF is significantly more performant than the two baselines.
The table also shows the results on the denoising task, and we observe that UNet-AF is significantly more equivariant than UNet-Ronneberger and UNet-Jin with similar values than for the circular deblurring task. In this setting however, the three models have comparable performance.

\subsubsection{Ablation on UNet-AF Fast} \label{subsec:ablation_unet_af_fast}

\Cref{tbl:abl_fast} shows the impact of the padding mode in filtered activations, and of the filters. In particular, switching from zero-padding to circular padding enables a gain in equivariance of more than~5 dB. If the filters are left unchanged however, the performance in terms of PSNR drops by~5 dB. Our proposed configuration obtains high performance, and equivariance.

\begin{table}[htbp]
    \centering
    \caption{
    \textbf{Performance on additional tasks.}
    We evaluate UNet-AF on valid deblurring, and Gaussian denoising.
    Best metric in \textbf{bold}, second best \underline{underlined}.
    Values: avg ± st.d. \\
    }
    \label{tbl:additional_tasks}
    \setlength{\tabcolsep}{6pt}
    \begin{tabular}{llcccc}
        \toprule
        Method & Task & PSNR $\uparrow$ & EQUIV $\uparrow$ \\
        \midrule
        UNet-Ronneberger & Valid deblurring & 31.67 ± 2.33 & 39.96 ± 2.71 \\
        UNet-Jin & & \underline{32.25 ± 2.41} & \underline{42.82 ± 2.96} \\
        UNet-AF  & & \textbf{32.30 ± 2.42} & \textbf{46.57 ± 2.58} \\
        \midrule
        Blurred images & & 26.85 ± 2.55 & N/A \\
        \midrule
        UNet-Ronneberger & Gaussian denoising & \textbf{29.70 ± 1.56} & 39.80 ± 1.31 \\
        UNet-Jin & & 29.58 ± 1.61 & \underline{40.64 ± 1.78} \\
        UNet-AF & & \underline{29.66 ± 1.57} & \textbf{78.96 ± 1.44} \\
        \midrule
        Noisy images & & 20.00 ± 0.03 & N/A \\
        \bottomrule
    \end{tabular}
\end{table}

\begin{table*}[tbhp]
    \centering
    \caption{
    \textbf{Ablation on UNet-AF Fast.}
    Best metric in \textbf{bold}, second best \underline{underlined}.
    Values: avg ± st.d. \\
    }
    \label{tbl:abl_fast}
    \setlength{\tabcolsep}{12pt}
    \begin{tabular}{llcccccc}
        \toprule
        Filters & Padding & PSNR $\uparrow$ & SSIM $\uparrow$ & LPIPS $\downarrow$ & EQUIV $\uparrow$ & STAB $\uparrow$ & FPS $\uparrow$ \\
        \midrule
        Kaiser & Zeros & \underline{32.41 ± 2.42} & 0.9404 ± 0.0209 & 0.0363 ± 0.0210 & 43.12 ± 1.89 & 22.30 & \underline{556} \\
        Kaiser & Circular & 26.79 ± 2.24 & 0.8315 ± 0.0367 & 0.2247 ± 0.0806 & \textbf{89.35 ± 2.35} & 13.45 & \textbf{571} \\
        Tuned & Zeros & \textbf{32.45 ± 2.44} & \textbf{0.9413 ± 0.0211} & \underline{0.0362 ± 0.0210} & 43.94 ± 2.06 & \underline{23.29} & 534 \\
        Tuned & Circular & 32.39 ± 2.40 & \underline{0.9408 ± 0.0207} & \textbf{0.0307 ± 0.0181} & \underline{49.36 ± 2.84} & \textbf{23.29} & 552 \\
        \bottomrule
    \end{tabular}
\end{table*}

\end{document}